\documentclass{article}
\PassOptionsToPackage{numbers, compress}{natbib}

\usepackage[final]{neurips_2023}

\usepackage[utf8]{inputenc} 
\usepackage[T1]{fontenc}    
\usepackage{hyperref}       
\usepackage{url}            
\usepackage{booktabs}       
\usepackage{amsfonts}       
\usepackage{nicefrac}       
\usepackage{microtype}      
\usepackage{xcolor}         
\usepackage{graphicx}
\usepackage{comment}
\usepackage{titlecaps}

\title{GameGPT: Multi-agent Collaborative Framework for Game Development}

\author{%
  Dake Chen \\
  AutoGame Research \\
  \texttt{dakechen@usc.edu} \\
  \And
  Haoyang Zhang \\
   AutoGame Research\\
  \texttt{17@autogame.ai}\\
    \AND
  Hanbin Wang \\
  X-Institute\\
  \texttt{wanghanbin@mails.x-institute.edu.cn}\\
  \AND
  Yunhao Huo \\
  University of Southern California \\
  \texttt{hhuo@usc.edu} \\
  \And
  Yuzhao Li \\
  AutoGame Research \\
  \texttt{ram@autogame.ai} \\
  \And
  Junjie Wang \\
  Tsinghua University  \\
  \texttt{wangjunjie@sz.tsinghua.edu.cn} \\
}

\begin{document}

\maketitle


\begin{abstract}

The large language model (LLM) based agents have demonstrated their capacity to automate and expedite software development processes.
In this paper, we focus on game development and propose a multi-agent collaborative framework, dubbed GameGPT, to automate game development.
While many studies have pinpointed hallucination as a primary roadblock for deploying LLMs in production, we identify another concern: redundancy. Our framework presents a series of methods to mitigate both concerns.
These methods include dual collaboration and layered approaches with several in-house lexicons, to mitigate the hallucination and redundancy in the planning, task identification, and implementation phases. Furthermore, a decoupling approach is also introduced to achieve code generation with better precision. 
\begin{figure}[t]
    \centering
    \includegraphics[width=1
    \columnwidth]{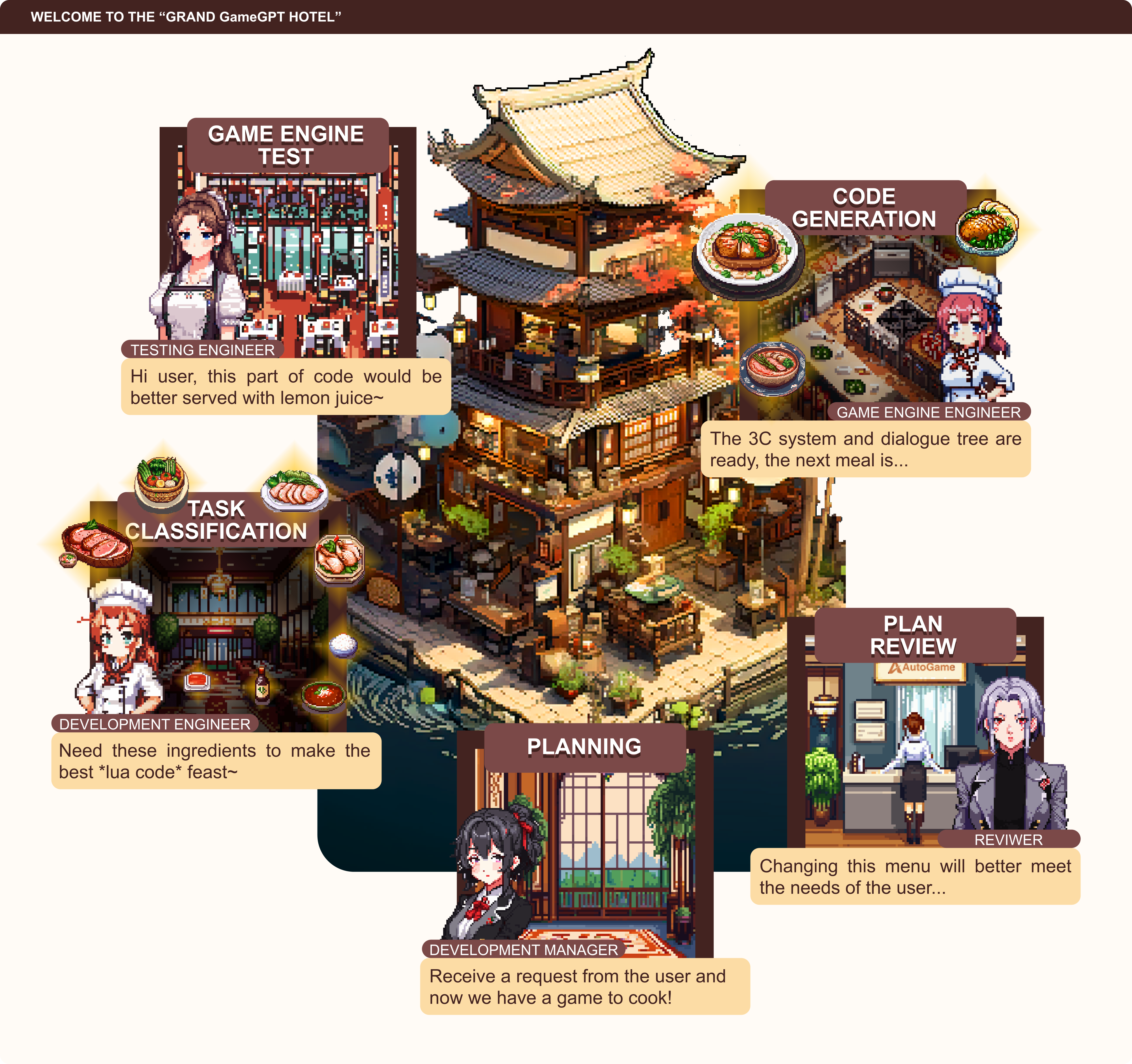}
    \label{fig:overview}
\end{figure}
\end{abstract}

\section{Introduction}

Artificial intelligence's applications in game development can be traced back to classic games such as \textit{Starcraft} and \textit{Diablo}~\cite{laird2001human, perez2016general,leverageml}. Developers have consistently required AI systems for crafting interactive virtual worlds and characters. These systems have become standard in the development of such interactive platforms. Early game AI research emphasizes controlling non-player characters (NPCs) and pathfinding~\cite{berner2019dota}. 
With the advancement of natural language processing (NLP), some pioneering works that focus on generating levels using the deep learning technique have emerged. A representative is MarioGPT~\cite{sudhakaran2023mariogpt}, which successfully generates levels in \textit{Super Mario Bros} by fine-tuning GPT2~\cite{radford2019language}.

Recently, transformer-based large language models (LLMs) have achieved substantial advancements, making notable strides in both natural language processing and computer vision~\cite{vaswani2017attention, radford2018improving, devlin2018bert, radford2019language, brown2020language, touvron2023llama, chiang2023vicuna}. The training of LLMs is a multi-phase process. The initial phase involves training these models on extensive corpora, fostering the acquisition of fundamental language capabilities. In the subsequent phase, which is of considerable significance, the models are fine-tuned via data for a diverse of NLP tasks that are delineated through instructions~\cite{hendrycks2020measuring, chung2022scaling, wang2022self, xu2023wizardlm}. This instruction tuning enhances the models' ability to generalize across a wide range of applications, leading to noteworthy zero-shot performance on unseen tasks. Lastly, the reinforcement learning from human feedback (RLHF) phase guarantees the models' structural integrity and reliability~\cite{ouyang2022training}. More importantly, this phase also grants the model the capacity to generate content that emulates human style, thereby enhancing its versatility as an agent.

Moreover, the advancement of LLMs has catalyzed the utilization of agents in automating software development processes. Various studies have explored the deployment of a single LLM-based agent to perform diverse tasks. AutoGPT, for instance, employs an LLM agent to tackle real-world decision-making tasks~\cite{yang2023auto}, while HuggingGPT employs a single LLM as a controller to orchestrate the completion of complicated AI tasks~\cite{shen2023hugginggpt}. Despite these approaches relying on a sole LLM agent, they incorporate reviewer roles to refine decision-making. In AutoGPT, a secondary opinion is obtained from a supervised learner to augment performance, and HuggingGPT integrates GPT-4 as a critic to evaluate decision accuracy.
Furthermore, multiple works utilize multiple agents in their frameworks to make LLM competent for complex tasks~\cite{qian2023communicative, hong2023metagpt, park2023generative, li2023camel, gong2023mindagent}. MetaGPT~\cite{hong2023metagpt} introduces a multi-agent framework, which can be used for automating the development of various software. CHATDEV~\cite{qian2023communicative} presents a novel software development framework that harnesses agents to enhance collaboration among the various roles involved in the software development process.

When employing LLM agents for automated software development, these studies encounter inherent limitations associated with LLMs, notably the issue of hallucination. This challenge manifests particularly during the planning and code generation phases~\cite{shen2023hugginggpt, qian2023communicative}. Distinct from generic software development, the game development industry operates under a stringent demand to keep up with the trends, necessitating heightened precision and conciseness throughout the development process for optimal efficiency. Moreover, tuning and employing one single LLM to serve the whole development cycle of game development without hallucination and high precision is impractical and costly. As a result, the framework requires multiple critic and reviewer roles to effectively mitigate the hallucinatory tendencies inherent in language models. Furthermore, in the context of game development, we identify an additional limitation of LLMs, that of redundancy. Particularly in the game development domain, LLMs can generate unnecessary and uninformative tasks or code snippets. 

To effectively address both hallucination and redundancy, the GameGPT framework strategically employs a combination of approaches, including dual collaboration, instruction tuning through in-house lexicons, and code decoupling. Notably, dual collaboration involves the interaction between LLMs and small expert deep learning models, alongside the collaborative engagement between execution roles and review roles. These synergistic have empirically demonstrated their effectiveness for mitigating both hallucination and redundancy within the framework.
In summary, our contributions are as follows:
\begin{itemize}
    \item We introduce an innovative multi-agent framework tailored to facilitate automated game development.
    \item Beyond hallucination, we identify the issue of redundancy inherent to LLM-based agents in the context of game development.
    \item To address the hallucination and redundancy concerns  of LLMs within game development, several mitigations including dual collaboration and code decoupling are proposed.
    \item Empirical results demonstrate the GameGPT's capability in effective decision-making and decision-rectifying throughout the game development process.
\end{itemize}

\section{GameGPT}
\subsection{Overview}

\begin{figure}[t]
    \centering
    \includegraphics[width=1.1\columnwidth]{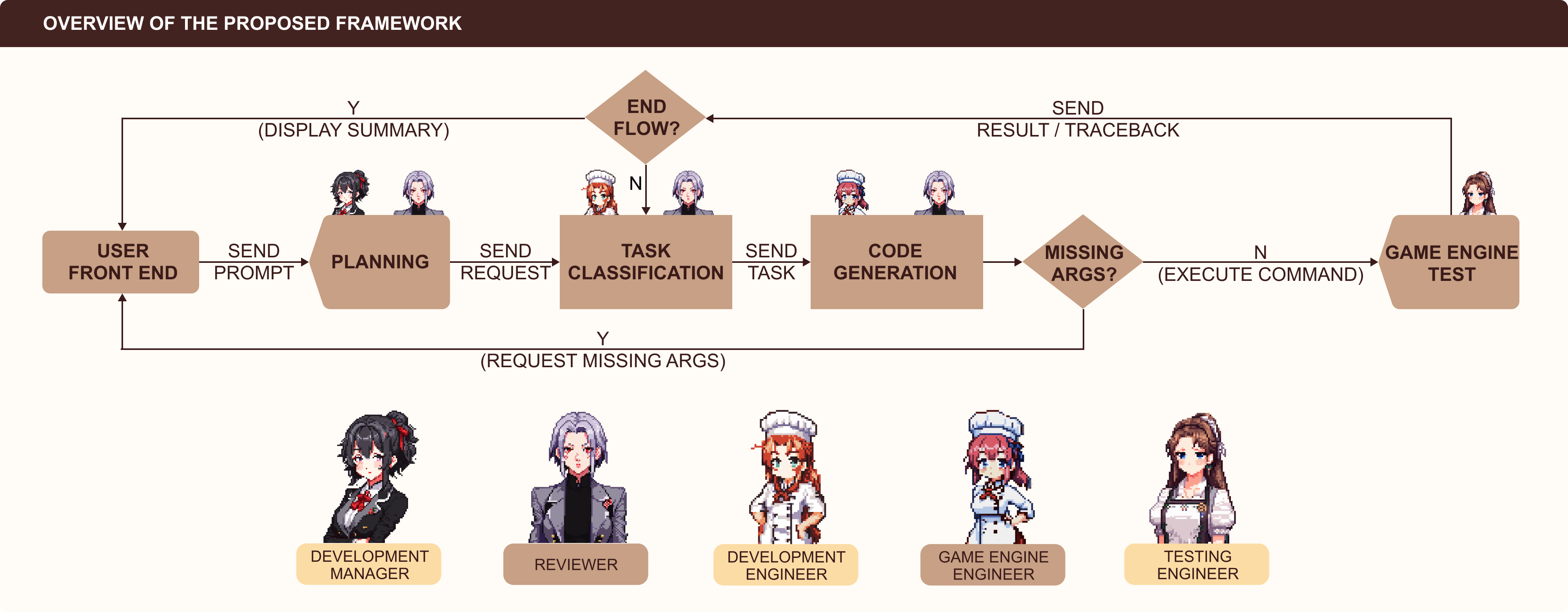}
    \caption{Overview of the proposed framework}
    \label{fig:overview}
\end{figure}

The GameGPT framework is designed as a specialized multi-agent system for game development. To address the  limitations of the LLM and the temporal constraints of game development, we integrate multiple agents with distinct roles into the framework. This integration aims to enhance precision and scalability. The scalability aspect of GameGPT offers the potential to create games of medium to large sizes. Moreover, GameGPT operates in a collaborative manner, exhibiting a dual collaboration approach. Firstly, it involves cooperation between the LLMs and smaller expert models dedicated to specific tasks, thereby enhancing the decision-making process. Secondly, collaboration occurs among agents assigned different roles, contributing to the decision-rectification and minimizing the hallucination of LLMs.

Figure~\ref{fig:overview} provides an overview of the proposed GameGPT framework, which operates through five distinct stages: game development planning, task classification, code generation, task execution, and result summarization. Upon receiving a client's request, the game development manager initiates the game development planning stage, resulting in the creation of a task list. Subsequently, game development engineers utilize smaller expert models to accurately determine the task type and its associated parameters. Following this, game engine engineers proceed to generate code and scripts in alignment with the designated game engineer. Throughout the initial three stages, three critics are incorporated to mitigate concerns related to hallucination and redundancy. Concluding these stages, the game engine testing engineer undertakes the execution of tasks and subsequently produces a comprehensive result summary.

\subsection{Multi-agent Framework}
In GameGPT, each agent maintains a private memory system and can access the shared public discussion to acquire the necessary information for guiding their decision-making process. For agent $i$ at time step $t$, this process can be formally represented as follows:
\begin{equation}
    \label{eq:decisioon}
    p_{\theta_i}(O_{it}|M_{it}, P_{t}),
\end{equation}
where $p_{\theta_i}$ corresponds to the LLM or an expert model associated with agent $i$, $O_{it}$ denotes the output or deliverable produced by the agent $i$ at time step $t$, and $M_{it}$ and $P_t$ refer to its private memory and the requisite public record up to time step $t$, respectively.

The presence of multiple agents with distinct roles is crucial in GameGPT due to the unique characteristics of the game development industry and the limitations of LLMs. Given that game development cycles typically span several months, relying on a solitary agent with comprehensive memory and contextual information would render language models, including LLMs, inefficient. This approach could also result in scalability challenges as projects become more complicated over time. Furthermore, considering the limitations on the number of tokens processed by LLMs, employing a solitary agent with an all-encompassing memory for large-scale game development projects is not pragmatic. Additionally, the inherent issues of hallucination and redundancy observed in LLMs underscore the significance of collaboration among multiple agents, particularly those with critic roles. This collaboration is significant in mitigating the challenges posed by LLM hallucination and redundancy. GameGPT utilizes a diverse set of roles to facilitate its operations, including responsibilities across the game development cycle. These roles comprise the game content designer, game development manager, plan reviewer, game development engineer, task reviewer, game engine engineer, code reviewer, and game engine testing engineer. Each of these roles works on distinct missions throughout the game development process.
\subsection{Game Development Planning}
The initial phase of GameGPT, upon receiving a user request, involves generating a task plan. This planning phase stands as one of the pivotal steps, greatly influencing the seamless progression of the entire development process. Orchestrated by an LLM-based game development manager, this phase entails proposing an initial plan that is subsequently disassembled into a list of tasks. Notably, due to the limitations inherent in LLMs, this initial plan often exhibits instances of hallucinations, which give rise to unexpected tasks, and include redundant tasks that are uninformative or unnecessary. To address these challenges, we present four strategies designed to alleviate these concerns. All four strategies are orthogonal to each other and can be layered to achieve better effectiveness.
\paragraph{Game Genre Classification with Template}
The first strategy is to perform classification for the incoming request, aimed at discerning the genre intended for the game. Presently, the GameGPT framework accommodates development for five distinct game genres, namely, [e.g., action, strategy, role-playing, simulation, and adventure]. For each genre, we provide a standardized plan template, guiding the game development manager in completing the template with relevant information. By adopting this approach, the frequency of redundant tasks is notably reduced, while simultaneously mitigating the likelihood of hallucination occurrences.
\paragraph{Plan Review}
The second strategy involves the engagement of a plan reviewer which is another LLM-based agent. This plan reviewer operates through strategically crafted prompts, facilitating a comprehensive review of the task plan. Its primary objective is to minimize occurrences of hallucination and redundancy. The plan reviewer assesses the plan and furnishes feedback, aiming to refine and enhance its precision, efficiency, and conciseness. The insights provided by the plan reviewer serve as input to the game development manager, empowering the shaping of a task plan that is notably more accurate and refined.
\paragraph{Instruction Tuning the Language Model for Planning}
The third approach aims to tune and tailor the LLM of the game development manager for game development planning through specialized instructions. This fine-tuning process endeavors to yield a plan that is both more accurate and concise. To facilitate this, we collect and consolidate an in-house dataset comprising numerous input-output pairs. While these pairs do not conform to a standardized format concerning length or structure, they uniformly revolve around requests for game development. The corresponding outputs are supplied by adept game development practitioners. By adopting this approach, we effectively bridge the gap between the LLM's general linguistic capabilities and its aptitude for game development planning.
\paragraph{User Presentation and Rectification}
The fourth and final strategy serves as a safety net within the planning phase. Throughout the planning process, the game development manager consistently shares interim outcomes with the users on the frontend interface, enabling them to remain informed about ongoing developments. To augment this, an interactive method is integrated, empowering users to actively review, rectify, and enhance the plan in accordance with their expectations. This approach safeguards alignment between the devised plan and the users' desires.

\subsection{Game Development Task Classification}
The process of task Classification within GameGPT demands a high accuracy in identifying both the task type and its corresponding arguments. Consequently, to ensure the accuracy of this phase, the agent of the game development engineer role is allocated. This role is supported by the utilization of two expert models, which collaboratively engage in task classification. This collaborative approach enhances the accuracy and effectiveness of task and argument identification.

\paragraph{Task Classifier and Task Argument Identifier}
To circumvent the hallucination of language models and enhance the accuracy of the task classification, we provide a list of possible types of tasks in game development. 
In order to perform the classification, a BERT model is employed to effectively categorize each  task. The BERT model has been trained with an in-house dataset. This dataset contains data entries uniquely tailored to the tasks of game development. The input is a task drawn from the predetermined list, while the output corresponds to the task's designated category.

Identifying the argument involves another LLM. The agent provides a template that corresponds to the identified task type and subsequently prompts the LLM to populate this template. The incorporation of the template can elevate the accuracy of argument identification and significantly reduce the hallucination.



\paragraph{Task Type and Argument Review  }
In this phase, a task reviewer reviewer agent is empolyed. It is prompted to double-confirm that the identified type and argument of each class are reasonable. The review process includes if the task type is in the predetermined range and is the best fit for the corresponding task. It also involves reviewing the argument list and see if it aligns with the task. In some scenarios where inferring the argument based on the contextual task information and the user's request is unfeasible, GameGPT adopts a proactive approach. The task reviewer engages the user by initiating a prompt on the frontend interface and requests additional information necessary for the argument. This interactive method ensures the completeness of argument details even in instances where automated inference falls short.
\paragraph{Task Dependency and Execution Sequence}
The task reviewer agent is also responsible for discerning task dependencies and constructing a directed acyclic graph that encapsulates these relationships. Subsequent to the establishment of this graph, a breadth-first search algorithm is employed to traverse it. The traversal yields a determined sequence for task execution. This process ensures an orderly and systematic execution of tasks in accordance with their dependencies, resulting in a coherent and structured development progression.

\subsection{Code generation}
Generating lengthy code scripts with an LLM inherently carries a greater risk of encountering hallucinations and redundancy. In response, we introduce a novel approach to decoupling the code for game design, simplifying the inference process for the LLM and consequently mitigating both hallucination and redundancy.

\paragraph{Decouple the Script for Game Development }
The hallucination and redundancy tend to be more frequent when generating lengthy scripts. In response, our proposed framework introduces a novel decoupling approach specifically designed for game development, aimed at separating the Lua script. To achieve this, we strategically divide the expected script into numerous code snippets of manageable length for LLM processing.
This decoupling approach significantly eases the work of LLM and thereby mitigates the occurrence of hallucination and redundancy.

\paragraph{Post Decoupling In-context Inference}
In~\cite{brown2020language}, an effective inference method called in-context-learning to mitigate hallucination is proposed.  In GameGPT,  we adopt a similar post-training strategy built upon our decoupling method. As our decoupling approach breaks down task-related code into smaller code snippets, we no longer depend on lengthy example scripts for inference. Instead, we incorporate multiple example snippets into the prompt, effectively reducing both hallucination and redundancy.

\paragraph{Candidate selection}
Moreover, another technique integrated into GameGPT to counteract hallucinations involves generating a set of $K$ code snippets for each task. These snippets are subsequently tested within a virtual environment and simultaneously presented to the user. Both the testing process and user feedback are leveraged to identify and eliminate problematic candidates, leaving only the most viable option for execution. This approach serves to further minimize the occurrence of hallucinations.

\paragraph{Instruction Tuning}
Furthermore, we have collected an in-house lexicon comprising an extensive repository of code snippets designed for game development. Each of these snippets is annotated by labelers, providing clear instructions that specify their intended purpose. This high-quality lexicon serves as a valuable resource for fine-tuning our model.
\paragraph{Code Review and Enhancement}
Following the code generation by the game engine engineer, a code reviewer agent is engaged to thoroughly review and inspect the codebase.  The code reviewer performs a comprehensive assessment, actively seeking out any instances of deviation from the original request or unintended hallucinations present within the code.  
Upon thorough scrutiny, the code reviewer not only flags potential discrepancies but also furnishes recommendations for refining the code, ultimately yielding a more reasonable version.
Subsequent to the review process, the revised code, along with the code reviewer's feedback, is shared with both the game engine engineer and the user through the frontend interface. If the user deems it necessary, they can provide suggestions for code revision directly via the frontend interface. These suggestions are relayed to the code reviewer, who subsequently assesses and incorporates them as appropriate, fostering a collaborative and iterative approach to code enhancement.

\subsection{Game Development Task Execution and Result Summary}
Once the code generation and enhancement are finished, the responsibility transitions to a game engine testing engineer, tasked with executing the generated tasks. During this phase, the testing engineer adheres to the execution sequence formulated in the preceding stage. The execution process involves sending the code of each individual task to the game engine. Subsequently, the testing engineer performs the execution and keeps track of the logs during the execution.
Upon the completion of all tasks specified in the execution sequence, the testing engineer consolidates all the logs generated throughout the execution process. This compilation results in the creation of a succinct and comprehensive summary, which is then presented to the user through the frontend interface.

Additionally, the testing engineer also identifies and reports any observed tracebacks during the execution. These tracebacks serve as critical indicators that may necessitate adjustments to the execution or code, enabling the refinement of the overall process and contributing to the generation of a polished end product.









{
\small
\bibliographystyle{IEEEtran}
\bibliography{bibliography}
}


\end{document}